\documentclass[journal]{vgtc}                     

\onlineid{1143}

\vgtccategory{Research}

\vgtcpapertype{algorithm/technique}

\title{Self-Supervised Continuous Colormap Recovery from a \\ \visedit{2D} Scalar Field Visualization without a Legend
}

\author{%
  \authororcid{Hongxu Liu}{0009-0003-6326-8848},
  \authororcid{Xinyu Chen}{0009-0001-3499-4889},
  \authororcid{Haoyang Zheng}{0009-0001-5784-3186}, 
  \authororcid{Manyi Li}{0000-0002-5251-0462},
  \authororcid{Zhenfan Liu}{0009-0005-9089-8321},
  \authororcid{Fumeng Yang}{0000-0002-8401-2580},\\
  \authororcid{Yunhai Wang}{0000-0003-0059-6580}, 
  \authororcid{Changhe Tu}{0000-0002-1231-3392},
  and 
  \authororcid{Qiong Zeng}{0000-0002-2827-8261}
}

\authorfooter{
  \item Hongxu Liu, Xinyu Chen, Haoyang Zhang, Manyi Li, Zhenfan Liu, Changhe Tu, and Qiong Zeng (corresponding author) are with Shandong University. 
  E-mails: \{liuhongxu\_0227, zhenghaoyang1229, liuzhenfannn\}@163.com, \\xinyu.chen0468@gmail.com, and $\{$manyili, chtu, qiong.zn$\}$@sdu.edu.cn. 
    \item Fumeng Yang is with University of Maryland. E-mail: fy@umd.edu.
  \item Yunhai Wang is with Renming University of China. \\ E-mail: cloudseawang@gmail.com.
  
}

\abstract{%
Recovering a \zq{continuous colormap} from a single \visedit{2D} scalar field visualization \manyi{can be quite} challenging, especially in the absence of a corresponding color legend. 
\zq{In this paper}, we propose a \zq{novel} colormap recovery \manyi{approach} \zq{that extracts the colormap \manyi{from} a color-encoded \visedit{2D} scalar field visualization by} simultaneously \zq{predicting the colormap and underlying data} using a \emph{decoupling-and-reconstruction} strategy. 
Our \manyi{approach} first separates the input visualization into colormap and data \zq{using} a decoupling module, then reconstructs the visualization with a differentiable color-mapping module. \zq{To guide this process, we design a reconstruction loss} between the input and reconstructed visualization\visedit{s}, \zq{which serves both as} a constraint to \zq{ensure} strong correlation between colormap and data during training, and \zq{as} a self-supervised optimizer for fine-tuning the predicted colormap of {unseen} visualizations during inferencing. 
To ensure smoothness and correct color ordering \zq{in} the extracted colormap, we introduce a compact colormap representation using cubic B-spline curves \visedit{and} an associated color order loss.
We \zq{evaluate our method}   quantitatively and qualitatively on a synthetic dataset and a \zq{collection} of \zq{real-world} visualizations from the VIS30K dataset~\cite{Chen21}.
\zq{Additionally, we} demonstrate \zq{its} utility \zq{in two prototype applications---colormap adjustment and colormap transfer---and explore its generalization  to  visualizations with color \zq{legends and ones encoded using discrete color palettes}.}
} 

\teaser{
\setlength{\belowcaptionskip}{-2.5mm}
  \centering
  \includegraphics[width=.95\linewidth]{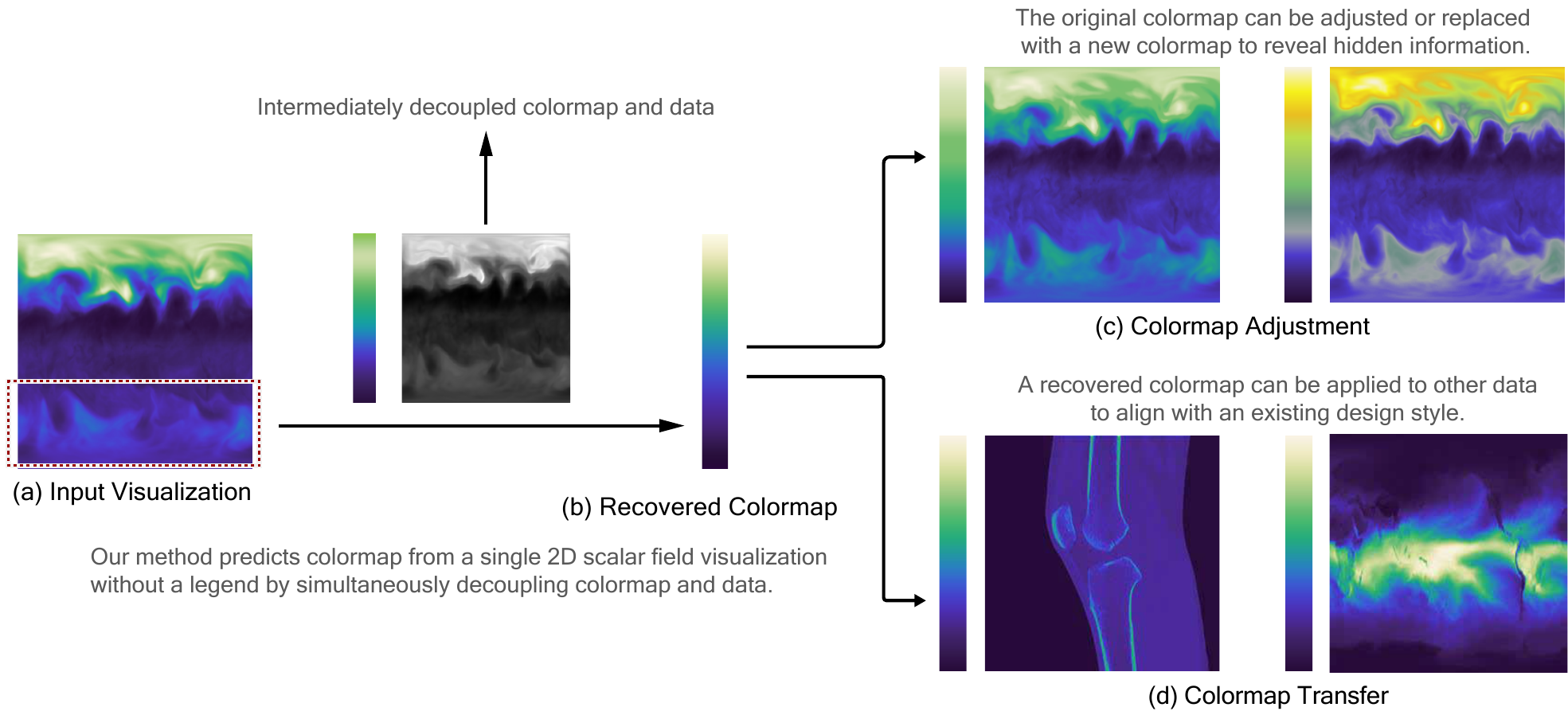}
  \caption{\textbf{Colormap recovery from a single visualization}. (a) shows a visualization with subtle \visedit{variations}\delete{changing patterns} in the \visedit{red dotted}\delete{highlighted} region. 
  (b) Our method successfully predicts the original colormap by simultaneously decoupling colormap and data embedded in the visualization. (c) Replacing the original colormap with two newly designed colormaps better reveals spatial variation\visedit{s} in the \visedit{highlighted region}\delete{data}. (d) \visedit{The} recovered colormap can be applied to new data for a consistent design style.}
  \label{fig:teaser}
}
\vspace{-3mm}

\graphicspath{{figs/}{figures/}{pictures/}{images/}{./}} 
\usepackage{tabu}                      
\usepackage{booktabs}                  
\usepackage{lipsum}                    
\usepackage{mwe}                       
\usepackage[pagebackref,bookmarks]{hyperref}
\usepackage{enumerate}
\usepackage{mathptmx}                  

\usepackage{textcomp}                  
\usepackage{times}                     
\usepackage{cite}                      

\usepackage{mathtools}
\usepackage{caption}

\usepackage{color}
\usepackage{enumitem}
\usepackage{amsmath}
\usepackage{amsfonts}
\usepackage{amssymb}
\usepackage{graphics}
\usepackage[utf8]{inputenc}
\usepackage{amsmath,bm}
\usepackage{multirow}
\usepackage{multicol}

\usepackage{subcaption}
\newif\ifnotes
\notesfalse 

\newif\ifedited

\def\fig{{Fig.}}

\def\sec{{Sec.}}

\newif\ifnotes
\notesfalse 

\newif\ifedited
\editedfalse 

\definecolor{fycolor}{rgb}{0.6, 0.4, 0.8}
\definecolor{editcolor}{rgb}{0, 0, 0}
\definecolor{viseditcolor}{rgb}{1, 0, 0}
\definecolor{black}{rgb}{0,0,0}

\newcommand{\edit}[1]{\color{editcolor}#1}
\newcommand{\visedit}[1]{\ifedited{\color{viseditcolor}#1}\else{\color{editcolor}#1}\fi}

\newcommand{\delete}[1]{\ifedited{\leavevmode\color{gray}\st{#1}}\fi}

\newcommand{\zq}[1]{{\color{black}#1}}

\newcommand{\manyi}[1]{{\color{black}#1}}

\def\etal{{\textit{et~al.}}}

\usepackage{soul}

\usepackage{afterpage}
\usepackage{xspace}
\usepackage[nameinlink,capitalise]{cleveref}
  \crefname{table}{Tab.}{Tabs.}
  \Crefname{table}{Table}{Tables}
  \crefname{section}{Sec.}{Secs.}
  \Crefname{Section}{Section}{Sections}
  \crefname{figure}{Fig.}{Figs.}
  \Crefname{Figure}{Fig.}{Figs.}

\begin{document}

\maketitle

\section{Introduction}
Color mapping is the process of assigning colors from a continuous colormap to data values through linear or non-linear mapping, creating a visualization where each color represents one or more data values.
In this process, the \zq{continuous} colormap is typically generated by \zq{linearly} or \zq{non-linearly} interpolating a sequence of colors\delete{, known as control points}. 
Color mapping is a powerful tool in scientific visualization (e.g., meteorology~\cite{Woodring2015} and fluid dynamics~\cite{Garth2007}), enabling researchers to effectively represent \visedit{2D} scalar fields and facilitate downstream analysis~\cite{Woodring2015,Dasgupta20}. 

The effectiveness of color mapping depends heavily on a number of design considerations, including data distribution, task requirements, and color choices~\cite{Tominski08,CHI2018Reda,Zeng22,Matee24}. 
Poorly designed colormaps can lead to misinterpretation~\cite{Crameri20}, such as the rainbow colormap exaggerating data variations due to its abrupt hue changes~\cite{Rog96}. This \delete{is a common }issue \visedit{is prevalent} in many \visedit{real-world} visualizations, \visedit{including those} found in published papers, online repositories, and informal reports~\cite{Poco18}\visedit{, where the original data are not provided}.
\visedit{For instance, in medical visualizations such as MRI imaging~\cite{Rog96}, a rainbow colormap can cause moderate values to appear as extreme due to perceptual discontinuities, misleading viewers and potentially affecting decision-making.}
\delete{An inappropriate colormap may obscure spatial variations in the data. For example}\visedit{Similarly, spatial patterns \delete{are missing from}\visedit{in the lower} \delete{the bottom }region of \visedit{\cref{fig:teaser}(a)} are obscured}\delete{ due to the poorly-designed colormap}.
\visedit{Moreover, many real-world visualizations fail to consider}\delete{, colormaps may not have considered} color vision deficiencies. \visedit{Colormaps that rely on red-green contrasts, for example, can render important features invisible to users with red-green color blindness}~\cite{Borland07,Nunez18,katrin22}\visedit{, thus limiting accessibility. These challenges highlight the need for effective colormap design and, more broadly, for techniques that can recover or replace poorly chosen colormaps in visualizations}. 
\delete{Given that many publicly available visualizations lack standardized design principles, improving colormap design is esssential to prevent misinterpretation in scientific communication, journalism, and public discourse. By enhancing colormap design, we can ensure visualizations are more accessible and interpretable, reducing the risk of miscommunication, especially in fields that rely on precise data interpretation.}

Colormap reverse engineering provides a potential solution to improve real-world visualizations~\cite{Manolis2011,Poco17,Poco18,Yuan22}. Its aim is to identify the colormap used and recover the mapping between colors and data values. This process typically relies on a color legend associated with the visualization to reconstruct the color-to-data mapping. 
The main challenge in these cases is to accurately detect colors and text within the legend using image processing and optical character recognition (OCR) techniques~\cite{Poco17,Poco18,Choi19}. 
However, many real-world visualizations, particularly those in books or on web pages,  lack a color legend~\cite{Yuan22}, making this reverse engineering even more difficult\delete{, if not impossible}.  
In such cases, accurate color extraction and correctly assigning data values to those colors are essential, \zq{making} this a fundamentally underconstrained problem\looseness=-1.

Several methods have been proposed to extract \zq{discrete color palettes} from visualizations without a legend. 
\zq{For instance, some} \zq{approaches} calculate categorical similarity or spatial distance between colors, and use heuristic priors or clustering algorithms to group similar colors~\cite{Yoo15,Chang15,Phan18}.
In a different approach, Yuan~\etal~\cite{Yuan22} introduced a convolutional neural network that extracts \zq{continuous} colormaps under the supervision of paired colormap and visualization examples. While it incorporates post-processing techniques \zq{such as Laplacian smoothing} to refine extracted \zq{colormaps, such steps} may introduce inaccurac\zq{ies} or noise. 
Additionally, Yuan~\etal's approach struggles with \zq{out-of-distribution} visualizations---ones with color distributions different from the training data, potentially resulting in \zq{incorrect or artificial} colors.

\visedit{Different from previous methods, w}e propose \visedit{an end-to-end} \textbf{self-supervised colormap recovery network} that directly learns \zq{continuous} colormaps from single visualizations without an accompanying legend. Our approach leverages the intrinsic relationship between visualizations, colormaps, and data, following a \emph{decoupling-and-reconstruction} strategy.
For simplicity and applicability, we focus on linear mapping between colormaps and data among \zq{different mapping strategies~\cite{Eisemann11,Thompson13,Tominski2020}}.
Our network operates in two stages.  
First, \textbf{an auto-encoder} decomposes the input visualization into a decoupled colormap and data. 
Second, \textbf{a differentiable colormapping module} reconstructs the visualization using the recovered colormap and data. 
\visedit{The network is trained in a supervised manner, where the training data consists of pairs of visualizations, their corresponding colormap, and data.}
\delete{This network relies on a}\visedit{A} \emph{reconstruction loss} between the original and reconstructed visualizations \visedit{is used} to guide \visedit{the training optimization process. This also allows the network to fine-tune the recovered colormap during inferencing with the input of a single visualization}\delete{ a self-supervised optimization process for fine-tuning recovered colormaps from out-of-distribution visualizations}.
To guarantee colormap smoothness, we use a compact representation based on cubic B-spline curves~\cite{Pham90,Bujack18}. To preserve the color order, we also introduce a color order loss that penalizes random color ordering.
\visedit{By explicitly decoupling the colormap from the data and incorporating a }\delete{Combining explicitly decoupling the colormap and data with the }self-supervised fine-tuning strategy, our method achieves robust colormap recovery even for out-of-distribution visualizations.

We compare our method to the state-of-the-art colormap recovery method~\cite{Yuan22}  quantitatively and through a perceptual study, based on synthetic and real-world visualizations. We generate the synthetic visualizations using colormaps from various sources~\cite{CHI2018Reda,Reda20,nardini2020a,Yuan22} to encode data from mathematical functions~\cite{nardini2020a} and the ECMWF public datasets~\cite{ECMWF}. 
The real-world visualizations came from the VIS30K dataset~\cite{Chen21}. 
We also conduct an ablation study to verify the importance of our core \emph{decoupling-and-reconstruction} components, including the cubic B-spline colormap representation, colormap recovery loss functions, and the fine-tuning procedure during inference. Finally, we show how to \zq{adapt our approach to general visualizations with legends or discrete color palettes, and} present the practical value of our method through \manyi{two} prototype applications: {colormap adjustment} and {colormap transfer}\looseness=-1.
Our main contributions are as follows:
\begin{itemize}[noitemsep]
  \item We propose \textbf{a decoupling-and-reconstruction network} using two design considerations extracted from the literature; our network supports self-supervised fine-tuning for colormap recovery from single visualization images lacking a legend\looseness=-1.   
  \item We introduce \textbf{a compact colormap representation} using cubic B-spline curves and a color order loss to ensure a visually smooth colormap with appropriate ordering. The use of cubic B-spline curves results in smoother transitions between colors, achieving at least a 29.6\% lower mean squared error (MSE) compared to traditional linear colormap representations.
  \item We show \textbf{the effectiveness} of our proposed network. Our model outperforms baseline approaches with an MSE improvement of at least \visedit{19.1}\%; it received a 26.5\%  higher rating in a perceptual study concerning the \visedit{visual quality of} \delete{similarity between the }recovered colormaps\delete{ and the original one}. 
  \item We demonstrate the practical usage of our approach in two \zq{application prototypes---\textbf{colormap adjustment} and \textbf{colormap transfer}---and extend it to accommodate general visualizations}. 
\end{itemize}

To facilitate future research, we have open-sourced our data, code, models, and interactive tool at \url{https://osf.io/gb2tx/?view_only=4d2a269b59bc4144b2806ec2d1a34e11}.
\section{Related Work}
Considerable efforts in the visualization community have been made to improve the effectiveness and efficiency of color design. 
For a complete investigation of color design in visualization, we refer readers to prior surveys~\cite{silva11,Bernard2015,Zhou16}.
In this section, we review colormap generation and optimization, colormap extraction, and image recoloring approaches.

\subsection{Colormap Generation \& Optimization}
Common color design considerations\delete{include}\visedit{, such as} color name distance~\cite{Heer12,Reda21} and perceptual uniformity~\cite{Bujack18}\delete{. 
The color design guidelines}\visedit{,} can either be directly modeled as colormap generation rules, or integrated with data distributions and visualization tasks~\cite{Mittelstadt15, Gramazio17,Nardini2019,Smart20,Lu21}. For example, Gramazio~\etal~\cite{Gramazio17} provided a color design tool, \emph{Colorgorical}, which incorporates user-assigned discriminability and preference into the process of color generation. Smart~\etal~\cite{Smart20} learned data-driven models to fit designer-crafted colormaps, and leveraged those models to generate an effective color palette from a single seed color. Lu~\etal~\cite{Lu21} integrated color generation and color assignment \delete{into a single optimization framework}for visualizing categorical data, \visedit{with}\delete{incorporating} color names and color discriminability as \visedit{optimization} constraints\delete{ to guide the optimization process}.

\begin{figure*}[t]
\centering
\includegraphics[width=.85\linewidth]{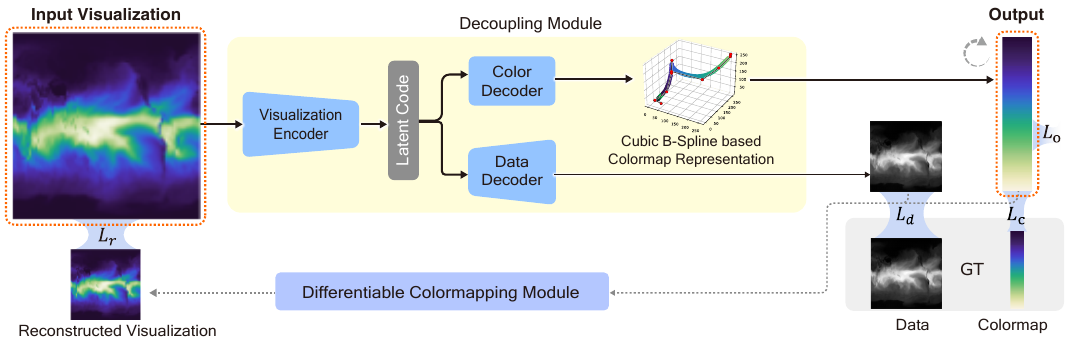}
\caption{\textbf{Overview of our colormap recovery network}. 
It consists of a \emph{decoupling module} and a \emph{differentiable colormapping module}. The decoupling module includes a visualization encoder to represent the input as a latent code, a color decoder to recover the colormap, and a data decoder to recover the underlying data. 
To ensure accurate colormap recovery, we employ four loss functions: a \textbf{color fidelity loss $\boldsymbol{L_{c}}$} to measure differences between the recovered and an original colormap, a \textbf{color order loss \boldsymbol{$L_{o}$}} to promote visually smooth color transitions, a \textbf{data loss} \boldsymbol{$L_{d}$} to measure differences between the recovered data and the ground-truth data, and a \textbf{reconstruction loss} \boldsymbol{$L_{r}$} to assess the similarity between  reconstructed and  input visualizations. 
During training, the network is trained on paired visualizations with known colormaps and data. 
During inferencing, the network operates on a single visualization without the latter information.}
\label{fig:architecture}
\vspace*{-5mm}
\end{figure*}

Another common challenge \visedit{in color design} is to improve a\delete{n input} colormap\delete{, which is unsuitable in some way}. 
Automatic optimization algorithms are proposed to enhance the original colormap to meet specific\delete{ color} design considerations related to colormap properties, data distributions, or task requirements~\cite{Tominski08,Thompson13,Nunez18,Nardini2021,Zeng22}. 
For example, 
Thompson~\etal~\cite{Thompson13} generated and adjusted colormaps to highlight prominent values based on heuristic data analysis rules. Nu\~{n}ez~\etal~\cite{Nunez18} enhanced colormaps while maintaining perceptual uniformity and linearity, especially for viewers with defective color vision. Tominski~\etal~\cite{Tominski08} and Zeng~\etal~\cite{Zeng22} adjusted control point positions within colormaps to better match the colormap to data histograms or spatial data variations. Additionally, Nardini~\etal~\cite{Nardini2021} presented a framework for local and global colormap optimization, adhering to colormap properties such as uniformity and color order. 

The colormaps resulting from these methods are used to \textit{forward} encode data in visualizations, typically presented with a legend. 
However, many \visedit{real-world}\delete{``in-the-wild''} visualizations lack such contextual information, making them difficult for users to interpret or reuse.
In this work, we reverse-engineer colormaps from \visedit{real-world}\delete{``in-the-wild''} visualizations without legends, and provide applications (e.g., \visedit{colormap adjustment and colormap transfer}\delete{visual recoloring and data views}) that enable users to interactively explore the underlying colormap\delete{ and the data} embedded within the visualization images.

\subsection{Colormap Extraction}
Extracting visual specifications from visualization images is a typical reverse engineering process, in which a visualization image serves as the input, and the extracted visual specifications and data values are the output. 
The key challenge here is to recognize chart elements (e.g., text, characters, and words in bar  charts~\cite{Manolis2011,Poco17,Jung17,Luo21,Mishra22,Masson23}) and infer their relationships. 
For example, Poco and Heer~\cite{Poco17} used text analysis to infer textual elements and deduce their correlations with graphical mark types. 
Luo~\etal~\cite{Luo21} unified deep neural networks and rule-based methods in a single framework, which includes key point and chart type detection, data range extraction, and chart object detection. 
However, these methods primarily address simple and discrete charts, without considering continuous\delete{ and complex} color mappings\delete{ as found} in scalar field visualizations.

One line of exploration is to incorporate additional information into visualization images to facilitate data recovery. For example, Zhang~\etal~\cite{Zhang21} embedded visual information into QR codes and placed them in unimportant areas of the visualization image. Fu~\etal~\cite{Fu21} encoded subtle chart information (e.g., coarse and fine marks) and concealed them using carefully designed opacity. 
However, these methods rely on altering the current design of visualizations and cannot be directly applied to existing real-world visualizations. 

The other line of work focuses on colormap recovery. 
Poco~\etal~\cite{Poco18} proposed a semi-automatic method to extract colormaps. It first detects and classifies the legend as either discrete or continuous, then uses \visedit{OCR}\delete{ORC} techniques to identify colors and extract the accompanying text. 
The extracted color and labeled text are further integrated to predict the colormap. 
However, this method assumes that the visualization includes an accurate legend, while many real-world visualizations lack well-defined legends, contain incomplete or misleading labels, or use inconsistent color mappings. In such cases, legend-based extraction might result in interpretation errors, highlighting the need for colormap recovery algorithms for visualizations without legends.

Our work focuses on inferring color mappings directly from a single visualization lacking a legend.
Yuan~\etal~\cite{Yuan22} shared similar goals to ours---recovering colormaps from single visualizations without legends. 
Their method extracts colormaps from a given visualization image using supervised deep neural networks trained on paired colormaps and visualizations.  
The process is followed by post-processing operations, including clustering for discrete colormaps and Laplacian smoothing for continuous ones, to enhance the quality of the extracted colormaps.
However, their method has two major limitations. First, it separates the deep colormap extraction from the post-processing steps, which can lead to inaccurate colors and the loss of mappings between the original colormap and data. 
Furthermore, the approach relies on paired datasets and lacks mechanisms to ensure robust performance on unseen visualizations with color distributions different from those in the training dataset (as discussed further\delete{ in} in~\cref{subsec:comp}). 
\vspace*{-1mm}
\subsection{Visualization Recoloring}
Research on visualization recolorization directly enhances input visualizations without first recovering colormaps~\cite{Elmqvist11,zhou2019,zhou2019a,zhou2020}. 
For example, Elmqvist~\etal~\cite{Elmqvist11} enhanced initial colors in user-defined regions of interest and applied a modified colormap to the data in those regions, but this method requires access to the original data.
Zhou~\etal~\cite{zhou2019a} sharpened color-mapped visualizations by compensating the power spectrum for viewing distance. Following this work, Zhou~\etal~\cite{zhou2020} further enhanced colors for high-dynamic range data by using tone mapping operators and overlaid glare for highlighting. The tone mapping operators preserved relationships between original data values and structures, while compressing the high-dynamic range into a suitable data range.
However, these methods enhance the original colormap without recovering the original colors, limiting their applicability in restyling and editing. 
It is worth noting that some image style transfer methods treat color recovery as an intermediate step, recoloring images based on the semantic relationships between the recovered color and reference palette\zq{s}~\cite{Chang15}. However, these methods mainly focus on preserving semantic correlations between images \zq{and may distort the underlying data in visualizations.} 

\section{Reverse-Colormapping Overview}
\label{sec:colormapping}
\subsection{Design Considerations}
\delete{Color mapping is the process of assigning colors from a continuous colormap to data values through linear or non-linear mapping, creating a visualization where each color represents one or more data values. In this process, the continuous }\visedit{The color mapping process typically assigns colors from a continuous colormap to data values, either linearly or non-linearly}~\cite{Tominski2020,Zeng22}. 
\visedit{Such} colormap\visedit{s are usually constructed}\delete{ is typically generated} by\delete{ linearly or non-linearly} interpolating a \visedit{set}\delete{sequence} of \visedit{predefined} colors, known as control points~\cite{Tominski2020,Zeng22}. 
To simplify the colormap recovery problem, we focus on \emph{linear mapping} between data values and colors\delete{ in continuous colormaps}.
Building on the color mapping foundations and insights of color design from the literature~\cite{Manolis2011,Poco17,Poco18,Bujack18,Nardini2019,Nardini2021,Zeng22}, we \visedit{define two core requirements (CR) that our framework must satisfy}\delete{propose two considerations} for effective colormap recovery.
\begin{itemize}[noitemsep, leftmargin=10pt]
    \item \textbf{CR1. High fidelity}. The recovered colormap and data should faithfully reflect the underlying color distribution, data patterns, and information encoded in the original visualization.
    \item \textbf{CR2. Visual clarity}. The recovered colormap should exhibit smooth color transitions and maintain consistent sorting order. This ensures clear visual communication and avoids introducing artifacts that could hinder data interpretation.
 \end{itemize}

\begin{figure*}[ht]
    \centering
    \includegraphics[width=.9\linewidth]{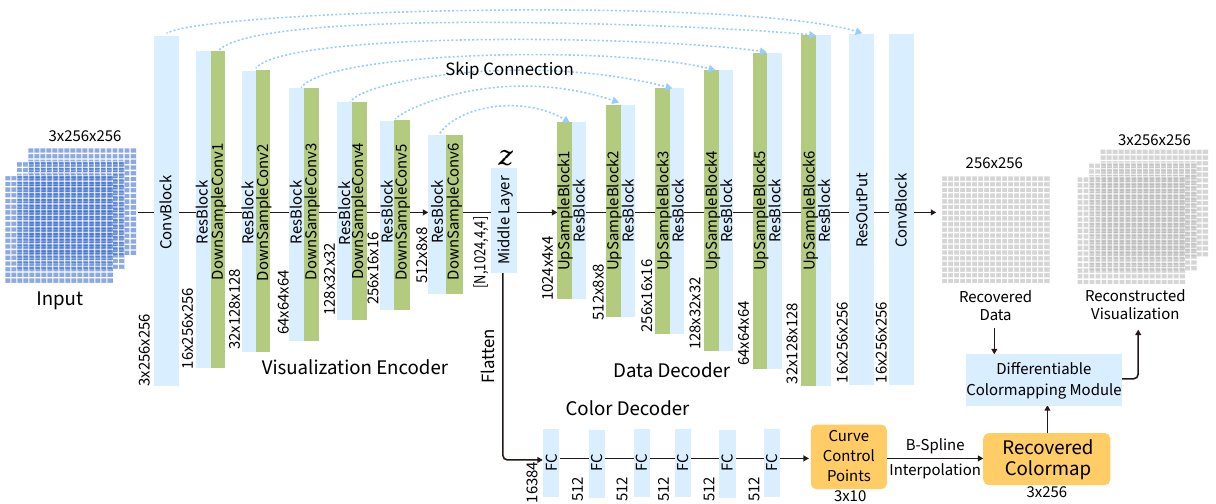}
    \caption{\textbf{Network architecture of our decoupling-and-reconstruction network}.The visualization encoder and data decoder are implemented as a convolutional U-Net with skip connections\visedit{~\cite{ronneberger2015u}}. The color decoder is an MLP network which outputs color curve control points, which are then transformed as a colormap and used to recover the visualization.}
    \label{fig:architecture1}
\vspace*{-5mm}
\end{figure*}
\vspace*{-4mm}
\subsection{Overview}
Motivated by the guidelines identified above, we propose to recover colormaps from single \visedit{2D} \zq{scalar field} visualizations by simultaneously decoupling the colormap and data, ensuring that the newly reconstructed visualization closely resembles the input.
\delete{To simplify the color recovery problem, we focus on \emph{linear mapping} between data values and colors in continuous colormaps. As such, t}\visedit{T}he colormap recovery procedure involves estimating a continuous colormap for unknown data based solely on a set of colors given in the input visualization.
This task is inherently underspecified, resulting in an infinite number of possible solutions. 
To address this issue, we propose to \emph{simultaneously decouple the colormap and the data}, while \emph{ensuring visual consistency} between the reconstructed and original visualizations using an end-to-end deep neural network (see \sec~\ref{sec:technical}). 
While our approach focuses on continuous colormaps, it can be easily adapted to discrete colormaps with minimal adjustments (see \sec~\ref{sec:general}). 

\section{Reverse-Colormapping Network}
\label{sec:technical}
Our network employs a \emph{decoupling-and-reconstruction} strategy  to separate the colormap information from the underlying data within a \visedit{2D} \zq{scalar field} visualization, which includes a \emph{decoupling module} (\sec~\ref{subsec:decoupling}) and a \emph{differentiable colormapping module} (\sec~\ref{subsec:reconstruction}). Figure~\ref{fig:architecture} shows the framework. 
\delete{We first formulate the problem, then detail the core components and loss functions of our network architecture, followed by the training and inference processes.}

\subsection{\visedit{Mathematical} Formulation}
\visedit{
Given an input visualization $\boldsymbol{I}\in\mathbb{R}^{M\times H\times W }$ (where $M$ is the number of visualization channels, $H$ is the height, and $W$ is the width), the goal of the colormap recovery network is to recover both the original colormap $\Xi\in\mathbb{R}^{n+1}$ (where $n+1$ is the number of control points in the colormap) and the underlying data $\boldsymbol{D}\in\mathbb{R}^{H\times W}$. 
The input visualization $\boldsymbol{I}$ is passed through the visualization encoder to produce a latent code $\boldsymbol{z}$, which is then fed into two separate decoder branches\visedit{, satisfying} \textbf{CR1}. 
The color decoder decouples $\boldsymbol{z}$ and recovers a set of control points $\boldsymbol{\hat{C}}$, 
which are used to generate a smooth colormap $\hat{\Xi}$ with a cubic B-spline interpolating function $\boldsymbol{S}$ (in line with \textbf{CR2}). 
The data decoder decouples $\boldsymbol{z}$ to recover the underlying data $\boldsymbol{\hat{D}}$. 
Then, the recovered colormap $\hat{\Xi}$ and data $\boldsymbol{\hat{D}}$ are fed into the \emph{differentiable colormapping module} to reconstruct the new visualization $\boldsymbol{\hat{I}}$.
Following the design guidelines, we define the total loss $L$ as the sum of four loss components: a \emph{color fidelity loss} $L_{c}$ to measure differences between the recovered and original colormap, a \emph{color order loss} $L_{o}$ to penalize disorder in the colormap, a \emph{data loss} $L_{d}$ to encourage the recovered data to faithfully represent the underlying data, and a \emph{reconstruction loss} $L_{r}$ to measure the difference between the reconstructed and original visualizations. 

To achieve our goal, we formulate the training objective for the colormap recovery network as the following optimization function:   
\begin{equation}
    \hat{\theta} = \arg\min\limits_{\theta} {\mathop{\mathbb{E}}}{(L(I,\Xi, D; \hat{I},\hat{\Xi},\hat{D}))},
\end{equation}
where $\theta$ represents the parameters of the network, $\hat{\theta}$ indicates the optimal parameters. $\mathbb{E}$ denotes that the total loss $L$ is averaged over the entire training set. During inference, $\hat{\theta}$ are fine-tuned in a self-supervised manner to recover colormaps for single visualizations without access to the original colormap $\Xi$ and data $D$. 
}\delete{Specifically, the visualization encoder first analyzes the input visualization $\boldsymbol{I}$ and generates a latent code $\boldsymbol{z}$. 
The latent code is then fed into the two separate decoder branches (\textbf{CR1}).
The color decoder recovers colormap information from the latent code and outputs a set of control points $\boldsymbol{\hat{C}}$ using the multi-layer perceptron (MLP) structure. 
These control points are then used to generate a smooth colormap $\hat{\Xi}$ with a cubic B-spline interpolating function $\boldsymbol{S}$, while avoiding disordered color assignment by applying a color order loss, in line with \textbf{CR2}.
In a separate decoder branch, the data decoder uses a classic convolutional neural network to predict the underlying data $\boldsymbol{\hat{D}}$.
The differentiable colormapping module combines the recovered colormap $\hat{\Xi}$  and data $\boldsymbol{\hat{D}}$  to reconstruct a new visualization $\boldsymbol{\hat{I}}$. 
By comparing it to the original input visualization $\boldsymbol{I}$, the network targets consistency between the recovered colormap and data, as required by \textbf{CR1}, so that colormap values are correctly mapped to the corresponding data variations. 
}
\delete{To ensure accuracy in colormap recovery, we define four loss functions: a \emph{color fidelity loss} $L_{c}$ to measure differences between the recovered and original colormap, a \emph{color order loss} $L_{o}$ to penalize color disordering, a \emph{data loss} $L_{d}$ to encourage the recovered data to faithfully represent the underlying data, and a \emph{reconstruction loss} $L_{r}$ to assess the similarity between the reconstructed and the original visualization. 
During training, the network learns to recover colormaps using these four loss functions and paired visualizations with the original colormaps and data. 
During inference, we fine-tuned the network parameters in a self-supervised manner to better adapt to out-of-distribution visualizations without access to the original colormaps and data.}

\subsection{Architecture}
\label{sec:architecture}
\delete{We first describe how a continuous colormap is represented, then detail the decoupling module and differentiable colormapping module in our  network, shown in Figure.}

\subsubsection{Colormap Representation}
In our framework, a colormap $\Xi$ is defined as a sequence of \emph{control points} and an \emph{interpolating function}~\cite{Zeng22}.
Control points are represented as an array of $n+1$ color values in RGB color space\visedit{\footnote{We trained our network in both the RGB color space and the perceptually uniform CIELAB color space, but chose RGB due to its widespread use in deep neural networks and its superior performance in our task (see \cref{sec:evaluate}).}}: $\boldsymbol{C} =\{\boldsymbol{c_0}, \dots, \boldsymbol{c_n}\}$, with all color components normalized to the range $[0,1]$. 
\delete{Although we trained our network in both RGB color space and the perceptually uniform CIELAB color space, we chose to use RGB color space due to its wide applications in training deep neural networks and its superior performance in our task. A comparison of the two color spaces is shown in.}To ensure visually\delete{ uniform and} smooth color transitions between control points, we use the cubic B-spline interpolating function~\cite{Pham90,Bujack18} rather than traditional piecewise linear interpolation~\cite{Zeng22}.\delete{ Our network predicts 10 control points to represent the recovered colormap.} The cubic B-spline interpolating function is described as follows:
\begin{equation}
    \boldsymbol{S}(t) = \sum_{i=0}^{n}{N_{i,3}(t)\boldsymbol{c_i}}.
\end{equation}
Here, $n+1$ gives the number of control points \visedit{(in our case $n=9$)}, and $\boldsymbol{S}(t)$ indicates the interpolated colors at parametric position $t$, as determined by the colors of the control points and the standard B-spline basis function $N_{i,j}(t)$ (where $i$ and $j$ are indices). The values $t_i$ represent knot positions of a cubic B-spline curve, arranged in non-decreasing order from $i=0$ to $n+4$. 

\subsubsection{Decoupling Module}
\label{subsec:decoupling}
Our network architecture consists of a visualization encoder, a color decoder, and a data decoder. Below we describe the network details, illustrated in Fig.~\ref{fig:architecture1}.
\delete{The visualization encoder extracts the essential features of the input visualization image, represented as a latent code of dimension \texttt{[1024,4,4]}. 
The color decoder and data decoder take the latent code as input and output the recovered colormap and data respectively. }

\vspace{.5em}
\noindent{\textbf{Visualization Encoder}}.
The visualization encoder $E$ aims to extract a latent code $\boldsymbol{z}$ that captures the essential features of the input visualization $\boldsymbol{I}$.
It begins by applying a single convolution block (denoted \emph{ConvBlock} in \fig~\ref{fig:architecture1}) to extract low-level features from the original visualization.  
Then, six downsampling modules, \visedit{each including a residual block (denoted as \emph{ResBlock})} and a downsampling block implemented as convolution layers (denoted as \emph{DownSampleConv}), are sequentially applied to compress the feature map. 
The final component is an intermediate layer, which is composed of a single ResBlock.
Note that all convolution layers in the visualization encoder have  $3\times3$ kernels.

\vspace{.5em}
\noindent{\textbf{Color Decoder}}. 
The color decoder, implemented as a multi-layer perceptron (MLP), is designed to decode the latent code $\boldsymbol{z}$ into a visually smooth colormap.
\zq{This process consists of two main steps:}
\begin{itemize}
    \item \textbf{Generating control points}: 
The MLP outputs a set of control points\delete{ that define a continuous colormap} in  RGB color space. This is achieved through six blocks, each containing a fully connected layer, a batch normalization layer, and an activation layer. The default number of control points is set to 10, so the MLP produces a $10 \times 3$ array of  RGB values (10 control points, each with three color channels).
\item \textbf{B-Spline Interpolation}: A cubic B-spline interpolation function is applied to generate the recovered colormap $\hat{\Xi}$ by interpolating between the estimated control points.
\end{itemize}

\vspace{.5em}
\noindent{\textbf{Data Decoder}}. The data decoder aims to recover the underlying data from the latent code $\boldsymbol{z}$. This auxiliary information serves to enhance the quality of the extracted colormap.
The decoder consists of six upsampling modules, each containing an upsampling block (denoted as \emph{UpSampleBlock}) with a convolutional layer and a residual block with two stacked convolutional layers. The upsampling block increases the spatial dimension of the latent data representation, \visedit{incorporating skip connections from the visualization encoder.} 

\visedit{In summary, our network is based on the U-Net architecture~\cite{ronneberger2015u} with three key modifications: (1) decoupling color and data into separate decoder branches, (2) incorporating residual blocks~\cite{He2016} for better detail preservation, and (3) replacing max pooling with convolutional layers for downsampling to improve feature extraction.}\delete{Additionally, residual connections are established between the visualization encoder and the data decoder to facilitate the recovery of intricate details and patterns in the original data.}

\subsubsection{Differentiable Colormapping Module}
\label{subsec:reconstruction}
Ideally, if the recovered colormap and data are consistent, the reconstructed visualization should closely resemble the input visualization.
Based on this assumption, we define a \emph{differentiable colormapping module} that reconstructs a visualization from the recovered colormap and data, following the colormapping procedure outlined in \cref{sec:colormapping}.
Since the module is differentiable, we can iteratively fine-tune the recovered colormap and data through backpropagation~\cite{Rumelhart1986} during training or inference, guided by the similarity between the reconstructed visualization and the ground-truth. This adaptability allows the module to effectively perform self-supervised learning, especially for real-world visualizations where no paired colormap and data are available. 

\subsection{Loss Function}
Motivated by the design guidelines, \visedit{we introduce} four loss functions \delete{are used as constraints }to optimize our network. \visedit{The color fidelity loss, data fidelity loss, and reconstruction loss ensure that the recovered colormap and data preserve high fidelity to the original visualization. Additionally, the color order loss encourages smooth and perceptually ordered color transitions in the recovered colormap.}

\vspace{.5em}
\noindent{\textbf{Color Fidelity Loss}}.
The color fidelity loss encourages similarity between the recovered colormap and the ground truth (GT) colormap. \visedit{It}\delete{The color fidelity loss} is calculated by measuring the mean squared error (MSE) between the GT colors and the ones in the recovered colormap; the latter is generated using a cubic B-spline and the control points predicted by the color decoder (see \edit{Sec.}~\ref{sec:architecture}). 
\begin{equation}
    L_\mathrm{\emph{c}} = \frac{1}{m} \sum_{i=1}^{m} ||\boldsymbol{\Gamma}(i)-\boldsymbol{\hat{S}}(t(i))||,
\quad    t(i) =  \frac{i}{m}.
\end{equation}
$\boldsymbol{\Gamma}(i)$ denotes the $i$-th color in the GT colormap. $\boldsymbol{\hat{S}}$ represents the recovered colormap. Here, $t(i)$ represents the parametric knot position on the cubic B-spline curve corresponding to the $i$-th color index. $m$ signifies the number of colors sampled from the \delete{continuous }color curve, typically set to $256$ to indicate a colormap with $256$ colors. \visedit{$||\cdot||$ denotes the difference between two colors, calculated using the Euclidean distance.}

\begin{figure}[t]
\centering
\includegraphics[width=\linewidth]{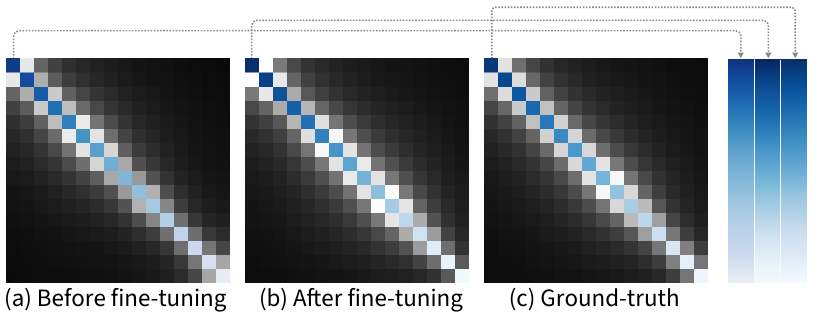}
\vspace*{-5mm}
\caption{\textbf{Color order}. (a--c) show the \emph{order ratio} between every distinct pair of colors, while the corresponding recovered colormaps are shown side-by-side on the right. (a) The recovered colormap before fine-tuning; black to white indicates increasing color order ratios. (b) The recovered colormap after fine-tuning: the lower right of the matrix shows similar color order values to those in the ground truth colormap in (c). }
\label{fig:order}
\vspace*{-5mm}
\end{figure}

\vspace{.5em}
\noindent{\textbf{Color Order Loss}}. 
To guarantee a natural color order, we introduce a color order loss\delete{ function}, inspired by \visedit{the} global legend-based color order~\cite{Bujack18,Nardini2021}. 
The core principle of the global legend-based order is to ensure that every color pair in a colormap is invertible and distinguishable\delete{ from each other}. 
We achieve this by calculating the \emph{order ratio} of color differences between each color pair in the recovered colormap to their \delete{corresponding }spatial distance along the colormap. 
A higher \emph{order ratio} signifies a better color order, where colors farther apart spatially exhibit \visedit{more}\delete{less} pronounced visual distinction.  
\delete{Since}\visedit{As} neural networks typically minimize loss functions, we use the \visedit{\emph{negative}} of the order ratio as our color order loss.
\begin{equation}
    L_\mathrm{\emph{o}} = -\min_{i\neq j\in{1,\dots,m}}\frac{||(\boldsymbol{\hat{S}}(t(i))-\boldsymbol{\hat{S}}(t(j) )||}{(i-j)^2},
\end{equation} 
where $i$ and $j$ represent the indices of two distinct colors within the recovered colormap. ~\Cref{fig:order} shows the \emph{order ratio} between every color pair in the recovered colormaps before and after fine-tuning.

\vspace{.5em}
\noindent{\textbf{Data Fidelity Loss}}. To encourage the recovered data to closely resemble the GT data, we introduce a data fidelity loss, which minimizes average differences between the scalar \delete{data }values predicted by our model and the corresponding values in the GT data. It is defined as follows:
\begin{equation}
    L_\mathrm{\emph{d}} = \frac{1}{W  H} \sum_{\visedit{h}=1}^{H}\sum_{\visedit{w}=1}^{W}(D(\visedit{h,w})) - \hat{D}(\visedit{h,w}))^2,
\end{equation} 
where $H$ and $W$ represent the height and width of the input. $D(\visedit{h,w})$ and $\hat{D}(\visedit{h,w})$ denote the normalized scalar value (ranging from 0 to 1) at position $(\visedit{h,w})$ in the GT data and the recovered data respectively. 
\begin{figure*}[ht]
\centering
    \includegraphics[width=.95\linewidth]{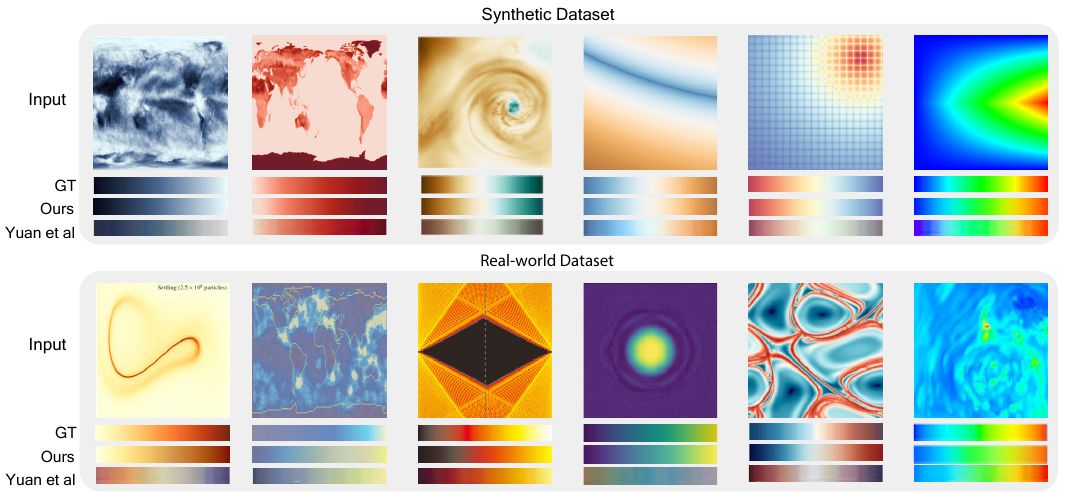}
\caption{\textbf{Comparison: our recovered colormaps and ones recovered by Yuan~\etal's method~\cite{Yuan22}}. 
\edit{Row 1: Colormaps recovered from synthetic data. Row 2: Colormaps recovered from a real-world data, with images screenshotted from~\cite{gunther2016backward,janicke2007multifield,bergner2006spectral,rapp2019void,Rogowitz96,liu2019scalable} in left-to-right order.} }
\label{fig:results}
\vspace*{-5mm}
\end{figure*}

\vspace{.5em}
\noindent{\textbf{Reconstruction Loss}}.
To guarantee faithful reconstruction of the color mapping, we introduce a reconstruction loss function. This loss minimizes average differences between corresponding pixels in the input and the reconstructed visualizations, defined as follows:
\begin{equation}
    L_\mathrm{\emph{r}} = \frac{1}{W  H} \sum_{\visedit{h}=1}^{H}\sum_{\visedit{w}=1}^{W}
    ||\boldsymbol{I}(\visedit{h,w}) - \hat{\boldsymbol{I}}(\visedit{h,w}))||,
\end{equation} 
where $\boldsymbol{I}(\visedit{h,w})$ indicates the RGB color at position $(\visedit{h,w})$. 

\subsection{Training and Inferencing}
\label{subsec:inference}
\subsubsection{Training}
During training, we combine the four proposed loss functions into a single \visedit{total loss}\delete{ objective function} using a linear weighted sum, allowing for straightforward gradient computation in the deep neural network:
\begin{equation}
    L =\alpha L_\mathrm{\emph{r}} +  \gamma  L_\mathrm{\emph{d}} +  \delta L_\mathrm{\emph{c}} + \eta L_\mathrm{\emph{o}}.
\end{equation} 
Here, $\alpha$, $\gamma$, $\delta$, and $\eta$ represent  balancing weights. \zq{Since all  loss functions are normalized to the range [0,1]}, we set these weights to 1 by default. 
The reconstruction loss $L_\mathrm{\emph{r}}$ and color order loss $L_\mathrm{\emph{o}}$ not only help the network infer hidden relationships between the colormap and data during training, but also serve as self-supervised constraints to fine-tune the trained parameters during inferencing. We further analyze the roles of these loss functions through ablation studies in \cref{subsec:ablation}. 

\subsubsection{Inferencing}
To enhance the generality of our method, we leverage the differentiable colormapping module and the color order loss to fine-tune the trained model in a self-supervised manner, without requiring paired original colormaps and data.
Specifically, we first use the trained model to generate an initial colormap and data directly from the input visualization.
\zq{In each subsequent iteration, we compute the reconstruction loss between the newly recovered colormap and the directly learned colormap (which replaces the original colormap in the loss function $L_\mathrm{\emph{c}}$), as well as the reconstruction loss between the newly recovered data and the directly learned data (which replaces the original data in the loss function $L_\mathrm{\emph{d}}$)}.
Using the newly recovered colormap and data at each iteration, we generate a corresponding reconstructed visualization and compute the reconstruction loss relative to the input visualization. Additionally, the color order loss is enforced to prevent arbitrary color arrangements.
Finally, the parameters of the trained model are fine-tuned by iteratively optimizing the recovered colormap and data to minimize both the\delete{ color fidelity loss,} color order loss\delete{, data fidelity loss,} and reconstruction loss.

\begin{table*}[t!]
\renewcommand\arraystretch{1.05}
\centering
\caption{\textbf{Quantitative comparison of recovered colormap quality} for our method and that of Yuan et al.~\cite{Yuan22}, using standard MSE, PSNR, and SSIM metrics. Values reported represent averages over all tested colormaps\visedit{, with the best-performing score highlighted in bold}.}
\label{tab:comparison}
\begin{tabular}{p{.13\textwidth}<{\centering}|p{.16\textwidth}<{\raggedleft}|p{.08\textwidth}<{\raggedleft}|p{.08\textwidth}<{\raggedleft}|p{.08\textwidth}<{\raggedleft}|p{.08\textwidth}<{\raggedleft}|p{.08\textwidth}<{\raggedleft}|p{.08\textwidth}<{\raggedleft}}
\toprule[1pt]
                            Dataset& \multicolumn{1}{c|}{Method}&  \multicolumn{3}{c|}{Recovered Colormap (\textit{ignoring direction})} &  \multicolumn{3}{c}{Recovered Colormap (\emph{considering direction})}\\ 
                            & &  MSE$\downarrow$ & PSNR$\uparrow$  & SSIM$\uparrow$  & MSE$\downarrow$ & PSNR$\uparrow$  & SSIM$\uparrow$ \\
                            \hline
                            \hline
\multirow{3}{*}{Synthetic Dataset}  
                    &   Ours (RGB)  & \textbf{0.001}& \textbf{34.033} & \textbf{0.961}&
\textbf{0.003}&	\textbf{33.739}&	\textbf{0.957}\\
 &   Ours  (LAB) & 0.015& 26.456 & 0.907&
0.026&	25.527&	0.887  \\
&   Yuan~\etal  & 0.011& 23.160&0.905 &  0.012	& 23.090 &	0.904\\

  \hline

\multirow{3}{*}{Real-world Dataset}  
&  Ours (RGB)  & \textbf{0.038} & \textbf{17.205} & \textbf{0.801} &0.071&	\textbf{14.910}&	\textbf{0.751}\\
&  Ours (LAB)& 0.055 & 14.224 & 0.748 &0.080&	12.762&	0.709\\
&   Yuan~\etal   & 0.047& 15.034&0.759 & \textbf{0.070}	&13.308 &	0.722 \\

                            \hline
\bottomrule[1pt]
\end{tabular}
  \vspace*{-3mm}
\end{table*}

\section{Evaluation}
\label{sec:evaluate}
We evaluate the effectiveness and utility of our method through the following experiments: (1) comparing to Yuan~\etal's method~\cite{Yuan22} on two datasets and three metrics, (2)  measuring the perceived quality of our recovered colormaps, and (3) conducting an ablation study to analyze the importance of individual components of our network architecture. Detailed results can be found in our supplementary materials.
\subsection{Implementation}
We implemented our network in PyTorch and trained it on an NVIDIA 4090D GPU for 50,000 iterations (about \visedit{6.5}\delete{14} hours) with a batch size of 64 and a learning rate of 0.001.  The Adam optimizer~\cite{kingma15} with $\beta_{0}$ set to 0.5 and $\beta_{1}$ set to 0.999 was employed to optimize the network parameters during training.
During testing, we performed fine-tuning with a learning rate of $0.0001$ for each input visualization. \delete{It took approximately two minutes to fine-tune a single image with 1,000 iterations.}
\subsection{Datasets}
\subsubsection{Synthetic Dataset} 
We constructed a synthetic visualization dataset by mapping commonly used colormaps to existing open-sourced scalar data and various mathematical functions, using the criteria of covering a wide range of data distributions and color variations~\cite{nardini2020a}. 
For the scalar data, we used two sources: (i) 179 randomly selected \visedit{2D} scalar fields from the ECMWF dataset (\url{https://apps.ecmwf.int/datasets}), 
and (ii)  228 \visedit{2D} scalar fields generated using various mathematical functions (e.g., linear gradients or periodic functions)~\cite{nardini2020a}.
For the colormaps in the synthetic dataset, we collected a total of 179 continuous colormaps from two sources: the Python matplotlib library and academic research papers on color design~\cite{CHI2018Reda,Reda20,nardini2020a,Yuan22}. 
Our synthetic dataset comprises 72,853 visualization images (179 $\times$ 179 + 228 $\times$ 179)\visedit{, with their accompanying colormap and data}. We randomly split the dataset into a \emph{training set} ($90\%$) and a \emph{test set} ($10\%$)\delete{ for building and evaluating the models}. 

\subsubsection{Real-world Dataset} 
\delete{In addition to the synthetic data, }We \visedit{further} compiled a dataset of 100 paired real-world visualizations and their corresponding colormaps, manually extracted from existing academic publications using the VIS30K dataset~\cite{Chen21}. We refer to this as the real-world dataset.
Specifically, we first identified 2D scalar field visualizations within the VIS30K dataset. For each visualization, we then reviewed the corresponding academic paper to check for the presence of an accompanying legend. If a legend was provided, we manually captured the corresponding colormap via screenshot \zq{to serve as the original ground-truth colormap for the visualization}; otherwise, the visualization was \zq{discarded}. \visedit{Note that all real-world visualizations were used exclusively for evaluation purposes and not for training the network. The corresponding colormaps were used solely as ground-truth for conducting quantitative and qualitative comparisons.}

\subsection{Quantitative and Qualitative Comparison}
\label{subsec:comp}
We evaluate \visedit{the performance of} our method \visedit{both qualitatively and quantitatively by comparing it}\delete{'s performance by comparing it, qualitatively and using three quantitative metrics,} to the state-of-the-art deep colormap extraction method \visedit{proposed by} Yuan~\etal~\cite{Yuan22}.

\begin{table*}[t!]
\renewcommand\arraystretch{1.05}
\centering
\vspace{1em}
\caption{\textbf{Quantitative comparison of different ablated networks} using MSE, PSNR, and SSIM metrics to assess recovered colormaps. Values reported represent averages over all tested colormaps\visedit{, with the best-performing score highlighted in bold and the second-best in Italic}.} 
\label{tab:results}
\begin{tabular}{p{.13\textwidth}<{\centering}|p{.16\textwidth}<{\raggedleft}|p{.08\textwidth}<{\raggedleft}|p{.08\textwidth}<{\raggedleft}|p{.08\textwidth}<{\raggedleft}|p{.08\textwidth}<{\raggedleft}|p{.08\textwidth}<{\raggedleft}|p{.08\textwidth}<{\raggedleft}}
\toprule[1pt]
                           Dataset &  \multicolumn{1}{c|}{Method}&  \multicolumn{3}{c|}{Recovered Colormap (\textit{ignoring direction})} & \multicolumn{3}{c}{Recovered Colormap (\emph{considering direction})}\\ 
                            & & MSE$\downarrow$ & PSNR$\uparrow$  & SSIM$\uparrow$  &  MSE$\downarrow$ & PSNR$\uparrow$  & SSIM$\uparrow$\\ 
                            \hline
                            \hline
\multirow{6}{*}{Synthetic Dataset}  
                    &   Ours with Fine-Tuning  &
\textbf{0.001}&	\textit{34.033}&	\textbf{0.961}&		\textbf{0.003}&	\textit{33.739}&	\textbf{0.957}\\
 &   Ours w/o Fine-Tuning &
\textbf{0.001}&	\textbf{34.054}&	\textit{0.958}&	\textbf{0.003}&	\textbf{33.756}&	\textit{0.954}\\ \cline{2-8}
  &  w/o B-Spline &0.028	&21.075&	0.867&		\textit{0.036}&	20.588&	0.854\\
&   w/o Recon. Loss &\textit{0.025}	&21.107&	0.882&		0.038&	20.109&	0.861\\
&  w/o D\&C Loss &0.060&13.618&	0.782&	0.109&	11.196&	0.718\\
 &   w/o Order Loss &0.028 	&20.432 &0.874	 &0.041	 &19.502	 &0.852 \\
  \hline
  
\multirow{6}{*}{Real-world Dataset}  
&  Ours with Fine-Tuning&\textbf{0.038}&	\textbf{17.205}&	\textbf{0.801}&	\textbf{0.071}&	\textbf{14.910}&	\textbf{0.751}\\
 & Ours w/o Fine-Tuning &\textit{0.040}&	\textit{17.071} &	\textit{0.797}&	\textit{0.073}&	\textit{14.710}&	\textit{0.745}\\ \cline{2-8}
 &  w/o B-Spline  &0.054&	14.476	&0.755&	0.078&	13.179&	0.720\\
&  w/o Recon. Loss &0.061&	13.667	&0.751&	0.076&	12.815&	0.725\\
 &  w/o D\&C Loss &0.085&	11.711	& 0.709&	0.104 &	10.689&	0.680\\             
&  w/o Order Loss  & 0.054	&14.561 &	0.761 &	0.083 &	13.123 &0.723\\
                            \hline
\bottomrule[1pt] 
\end{tabular}
  \vspace*{-3mm}
\end{table*}

\subsubsection{Approach} 
Yuan~\etal~\cite{Yuan22} introduced a convolutional neural network for colormap extraction from single visualizations\visedit{, and they have open-sourced their code at~\url{https://github.com/yuanlinping/deep\_colormap\_extraction}. We consider this method the most comparable and reproducible baseline, as other inspiring prior work~\cite{Poco17,Poco18} differs in focus (e.g., chart with legend) and lack compatible pipelines for direct comparison.} To ensure a fair comparison, we trained Yuan~\etal's model using their publicly available code\delete{at} and our training data. 
Note that we directly compared the output of the models without further post-processing for both methods.

\subsubsection{Metrics} 
To assess the \visedit{performance}\delete{quality} of our \delete{colormap recovery} method, we employed three \delete{quantitative} metrics: \emph{mean squared error} (MSE), \emph{peak signal-to-noise ratio} (PSNR), and \emph{structure similarity index measure} (SSIM). 
These metrics compare the recovered \visedit{colormaps}\delete{results (estimated colormap and data, and their reconstructed visualization)} with the corresponding ground-truth\delete{ values}.
\begin{itemize}[noitemsep]
  \item \emph{MSE} measures the average squared difference between the corresponding pixels in the recovered colormap and the corresponding ground truth. Lower MSE indicates a better match between the recovered elements and ground truth.
  \item \emph{PSNR} measures the ratio between the maximum signal (e.g., the maximum color value in the colormap) and the noise introduced during the recovery process, as quantified by the MSE. Higher PSNR indicates better reconstruction quality with lower noise.
  \item \emph{SSIM} goes beyond pixel-wise differences by incorporating structural information into the evaluation. It considers luminance, contrast, and structural similarity between the recovered results and the ground truth. A value closer to one indicates a higher level of perceptual similarity.
\end{itemize}

\subsubsection{Qualitative Comparison}
Figure~\ref{fig:results} compares visual results of our colormap recovery method to Yuan~\etal's. 
As demonstrated across a variety of examples\delete{ presented}, our method \visedit{shows} superior \visedit{performance} in  \visedit{recovering}\delete{its recovery of} continuous colormaps.
\visedit{Moreover, the}\delete{Additionally, our} recovered colormaps closely resemble the ground truth, \visedit{highlighting} the accuracy and visual fidelity of our approach.
\visedit{Additional}\delete{Further} qualitative comparisons\visedit{--including colormaps before fine-tuning, intermediate reconstructed data, and reconstructed visualizations}--are available in \visedit{the}\delete{our} supplementary material.

\subsubsection{Quantitative Comparison}
Table~\ref{tab:comparison}  quantitatively compares our method to Yuan~\etal's approach~\cite{Yuan22} using the MSE, PSNR, and SSIM metrics. 
We tested both methods using $999$ randomly selected visualizations from the test subset of the synthetic dataset, and all  $100$ visualizations from the real-world dataset.  We evaluated the alternatives in \zq{two} ways: 
\begin{itemize}[noitemsep]
  \item  \emph{Recovered Colormap (ignoring direction)} measures the accuracy of the predicted colormap without considering the order of colors.
  
  \item \emph{Recovered Colormap (considering direction)} measures the accuracy of the predicted colormap considering the order directions.
\end{itemize}

Our model \visedit{outperforms}\delete{performs better than} Yuan~\etal's in recovering colormaps, whether the colormap direction is ignored or \visedit{not}\delete{considered}. 
This superiority holds true for both synthetic and real-world datasets in  all cases considered.
However, when ordering direction is considered, the  results for both models are worse than when the direction is ignored, indicating that both models may occasionally produce colormaps with reversed directionality. 
\visedit{Our method requires five additional hours of training compared to Yuan~\etal's method, using the same machine and number of iterations. The colormap recovery process takes about 0.36 seconds, with an additional one minute for fine-tuning. While this is longer than Yuan~\etal's, it remains acceptable for interactive exploration. Detailed timing records can be found in the supplementary material.
}

\subsection{Perceptual Study}
\subsubsection{Experiment Design} 
To assess the perceived authenticity of our recovered colormaps, we conducted a \zq{one-factor within-subject} perceptual study. \zq{The single factor in the experiment was the colormap recovery method (either ours or Yuan \etal's). 
Each participant evaluated colormaps from both methods shown with random orders. 
The goal was to determine how similar participants perceived the generated colormaps to the ground truth in terms of color distribution.}
We randomly selected $40$ visualizations from the synthetic dataset and $20$ from the real-world dataset, and recovered their colormaps using both methods. A complete list of the experimental stimuli is provided in the supplementary material.

\subsubsection{Task} 
Participants were presented with a set of $120$ trials. 
Each trial displayed a ground-truth colormap on the left, and on the right a colormap reconstructed using our method or Yuan~\etal's method.
Participants then completed a forced-choice questionnaire on a five-point Likert scale to rate the perceived color distribution similarity between the two
\begin{figure}[h!]
\centering
\includegraphics[width=.95\linewidth]{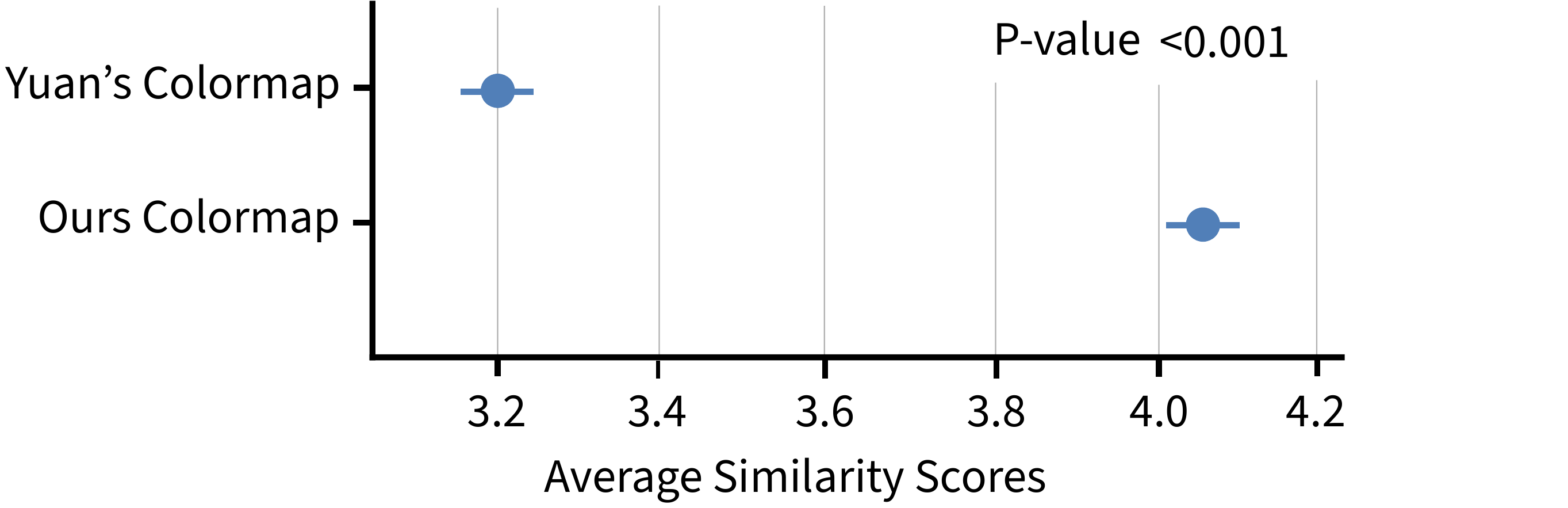}
 \vspace{-3mm}
\caption{\textbf{Perceptual study results}, \visedit{showing}\delete{giving} average similarity scores\delete{ assessed by participants. 
It shows the original average similarity scores} with $95\%$ CI\delete{, with} \visedit{and p-value} \delete{significant differences }between our \visedit{method}\delete{results} and the alternative\delete{s}.} 
\label{fig:userstudy}
\vspace*{-5mm}
\end{figure}
\begin{figure*}[t!]
    \centering
    \includegraphics[width=1\textwidth]{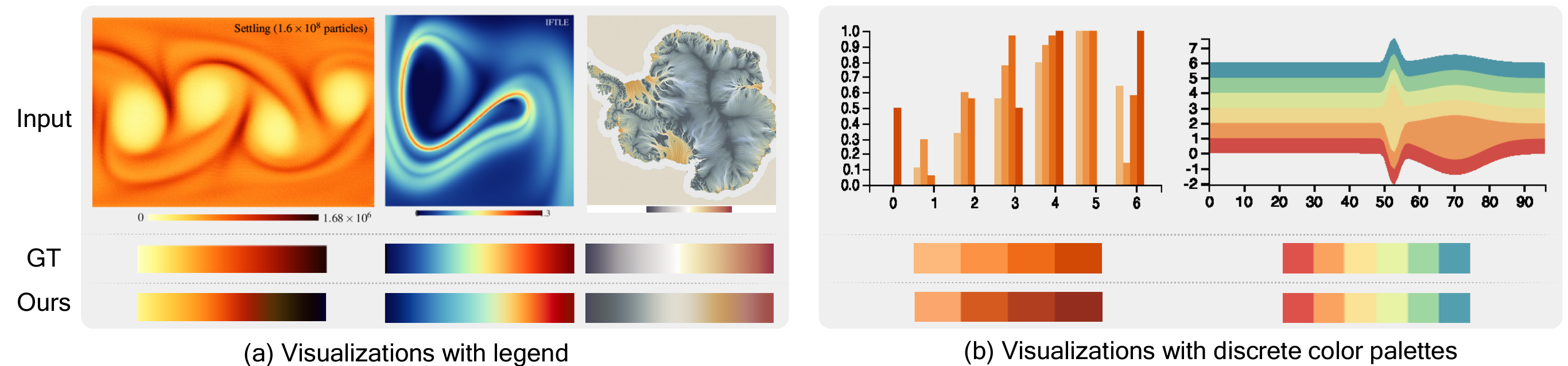}
    \caption{\textbf{\visedit{Adaptations}\delete{Extensions} of our method to general visualizations}. (a) Colormap recovery results from input visualizations with legends, taken from~\cite{abram2021antarctic,gunther2016backward}. (b) Visualizations from~\cite{Yuan22}, showing the recovery of discrete color palettes.}
    \label{fig:extention2}
 \vspace*{-3mm}
\end{figure*}
 colormaps. We asked the question ``How similarly do the colors appear to be distributed in the image on the right compared to the one on the left?'' with choices:
    (1) strongly dissimilar, 	 (2) slightly  dissimilar, 	(3) neutral, 
 (4) slightly similar, and (5) strongly similar.

\subsubsection{Procedure}
Our experiment \visedit{consists of three steps:}\delete{followed a common three-step structure, including} a training \visedit{phase}\delete{step}, the main experiment, and demographic information collection. 
The main experiment \visedit{included}\delete{consisted of} $120$ random trials for each participant, with $80$ trials using colormaps from the synthetic dataset and the remaining $40$ from the real-world dataset. 
We recorded their total response time and selected similarity scores from each trial for further analysis. 
On average, participants took 9 minutes to complete the experiment (\emph{minimum}: 3 minutes, \emph{maximum}: 17 minutes).

\subsubsection{Participants}
We recruited $40$ participants (30 males, 10 females) from the local university. To ensure data quality, participants were required to complete the experiment \delete{in front of }\visedit{using the same}\delete{a} computer screen (2048$\times$1080 resolution)\delete{.
Experimental sessions were conducted} in a consistently illuminated room.\delete{, with participants seated in front of a screen} 

\subsubsection{Results}
We collected $4,800$ valid results. 
Average similarity scores given by participants are summarized in \edit{Fig.}~\ref{fig:userstudy}. 
The plot shows the raw average user similarity scores along with $95\%$ confidence intervals~(CI).
Participants consistently determined our results to be more similar to the ground truth than Yuan~\etal's results. 
The differences between our results and Yuan~\etal's are statistically significant ($p<.001$), indicating a strong assessment of the superiority of our recovered colormaps.

\subsection{Ablation Study}
\label{subsec:ablation}
We conducted an ablation study to assess the importance of different components of our network by removing or replacing key components by alternative solutions.
We assessed our network as follows: 
\begin{itemize}[noitemsep]
    \item \textbf{Without fine-tuning during inference} (w/o Fine-Tuning): we directly \visedit{use} the trained model \visedit{to generate results} without \visedit{fine-tuning its parameters}\delete{applying the differentiable colormapping module and color order loss}.
     \item \textbf{Without using the cubic B-spline curve representation} (w/o B-Spline): we replaced the cubic B-spline curve representation with a traditional linear colormap representation~\cite{Zeng22}.
     \item \textbf{Without using the reconstruction loss} (w/o Recon. Loss): we trained the network without calculating the reconstruction loss between the recovered visualization and the input. 
    \item \textbf{Without data and color fidelity loss} (w/o D\&C Loss): we trained our network without incorporating the data fidelity loss and color fidelity loss\delete{ functions}.
     \item \textbf{Without color order loss} (w/o Order Loss): we omitted color order loss in the training stage.

\end{itemize}

This ablation study dissects the significance of various network components on our model's ability to recover hidden data and accurate color information from individual visualizations. Table~\ref{tab:results} lists \delete{the }result\delete{ing value}s of the various metrics. Further quantitative results, including different balancing weights and control points, are available in the supplementary material. Analyzing the results, we draw the following four conclusions:
\begin{itemize}[noitemsep]
    \item The absence of the component responsible for calculating data  and color fidelity loss (w/o D\&C Loss) significantly impacts performance, suggesting their critical roles in recovering colormaps. 
    \item Discarding  B-spline representation (w/o B-Spline), reconstruction loss (w/o Recon. Loss) and color order loss (w/o Order Loss) results in a general decrease in performance. 

    \item The real-world dataset benefitted more from the fine-tuning procedure. A possible reason is that synthetic datasets have simpler, more regular patterns, which the general model captures well. In contrast, the fine-tuned model, being more specialized, struggles with the simpler, different distribution of synthetic data. This suggests that fine-tuning improves performance on real-world visualizations but may introduce bias on simpler synthetic data.
\end{itemize}

\section{Applications and Discussions}
\subsection{Application Prototype}
\begin{figure}
    \centering
    \includegraphics[width=1\linewidth]{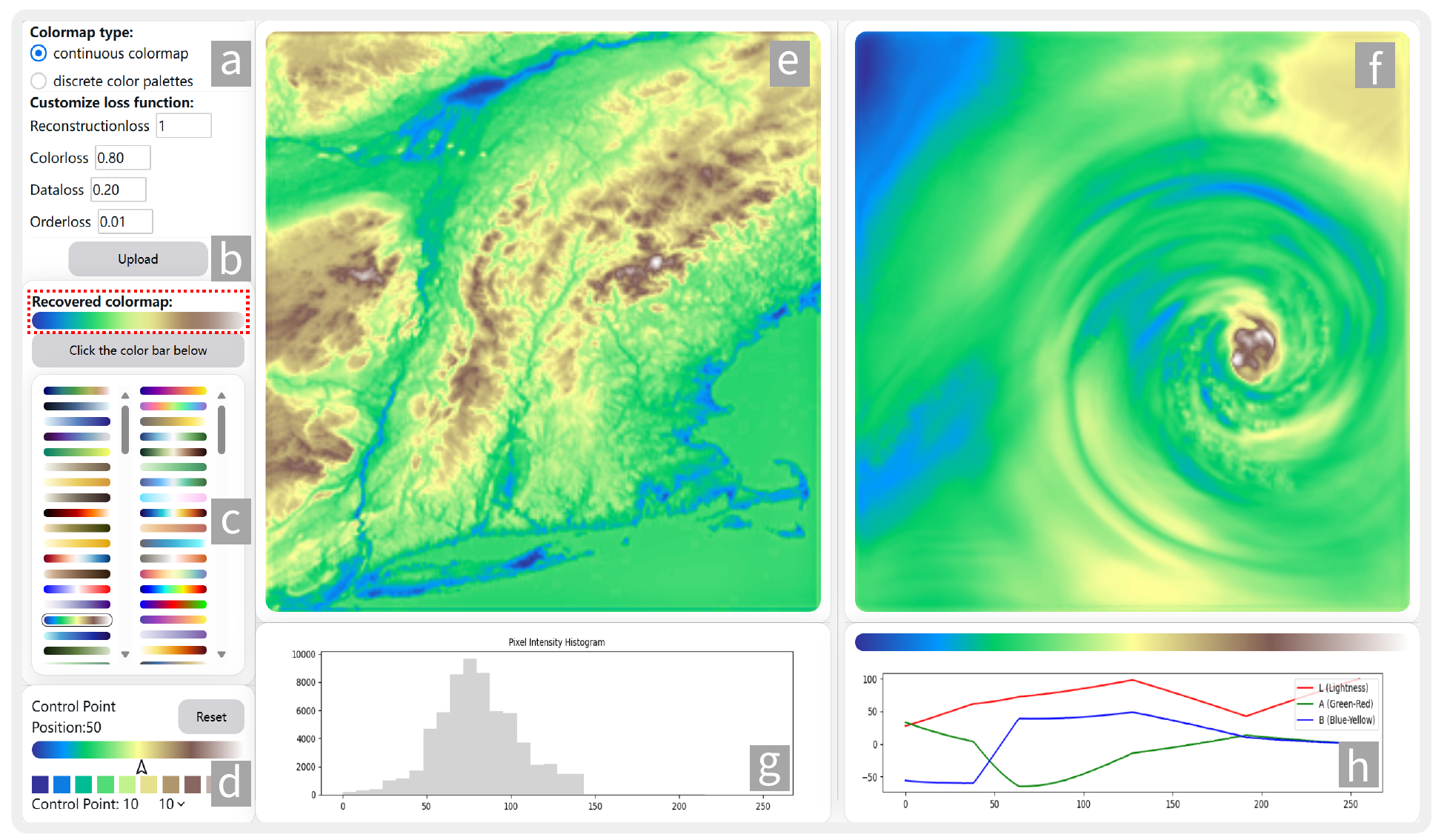}
    \caption{\textbf{Interface of our interactive tool.} The tool consists of eight functionalities, (a) parameter control, (b)  recovered colormap, (c) colormap selection, (d) colormap adjustment, (e) input visualization, (f) colormap transfer, (g) histogram distribution of the uploaded data by users, (h) color components of the recovered colormap.}
    \label{fig:tool}
    \vspace{-2em}
\end{figure}

Recovering hidden colormaps unlocks new possibilities for enhancing both machine and human understanding of static visualizations. 
This recovered information serves as a key for interactive exploration, empowering users to delve deeper into the data.
We demonstrate the utility of our method with two prototype applications: \textbf{colormap adjustment} and \textbf{colormap transfer}. 
\fig~\ref{fig:tool} shows the interface of our interactive tool; see our supplementary material, video, and accompanying tool for details on the interface and interactions.

\vspace{.5em}
\noindent\textbf{Colormap Adjustment}. 
\zq{Colormap adjustment involves either manually modifying the input colormap or applying a new colormap (e.g., a user-designed one) to the visualization.
\fig~\ref{fig:teaser}(c) re-renders the input visualization by (i) adjusting the colormap to allocate more colors to data ranges with higher density, and (ii) applying a carefully designed colormap to better reveal hidden spatial variations in the data}.

\noindent\textbf{Colormap Transfer}.
Colormap transfer applies the recovered colormap to new data, enabling users to leverage their preferred designs and to visually compare data. \fig~\ref{fig:teaser}(d) demonstrates two examples: applying the extracted colormap to a computed tomography dataset and a synthetic dataset simulating latitude changes.

\subsection{\visedit{Adaptation} to General Visualizations}
\label{sec:general}
\visedit{In this section, we demonstrate two straightforward ways to adapt our method to a broader range of visualizations.}

\subsubsection {Visualizations with legend}
\zq{The simplest way to generalize our approach to visualizations with a legend is to treat them as a whole within our network, i.e., by inputting the visualization and its paired colormap as a single image. \fig~\ref{fig:extention2}(a) presents three examples from existing 
work~\cite{abram2021antarctic,gunther2016backward}, in which corresponding color legends are provided. The color legends generated by our method closely match the original ones in the visualizations, demonstrating the effectiveness of our approach.}

\subsubsection{Visualizations with discrete color palettes} 
\zq{While our approach focuses on continuous colormap recovery, it can also handle discrete color palettes by sampling colors from the recovered continuous colormap. 
Two important factors need to \visedit{be} emphasized in this process: the number of colors in the discrete color palette and their corresponding values. 
Our basic idea is that meaningful discrete colors can naturally emerge from the distribution of data values. 
Therefore, we leverage DBSCAN~\cite{ester96} to cluster the estimated data from our approach, using the number of clusters as the number of colors in the discrete color palette. 
We then determine each color in the palette by selecting a color from our recovered continuous colormaps that corresponds to the average data value within each cluster.
\fig~\ref{fig:extention2}(b) presents two examples of visualizations with discrete color palettes, where the original input visualizations are sourced from Yuan~\etal's dataset}~\cite{Yuan22}. \visedit{Although the results shown in \fig~\ref{fig:extention2}(b) are promising, our method might struggle when dealing with discrete color palettes that violate the monotonic assumption underlying our color order loss.}

\subsection{Limitations} 
\begin{figure}[t]
    \centering
    \includegraphics[width=0.9\linewidth]{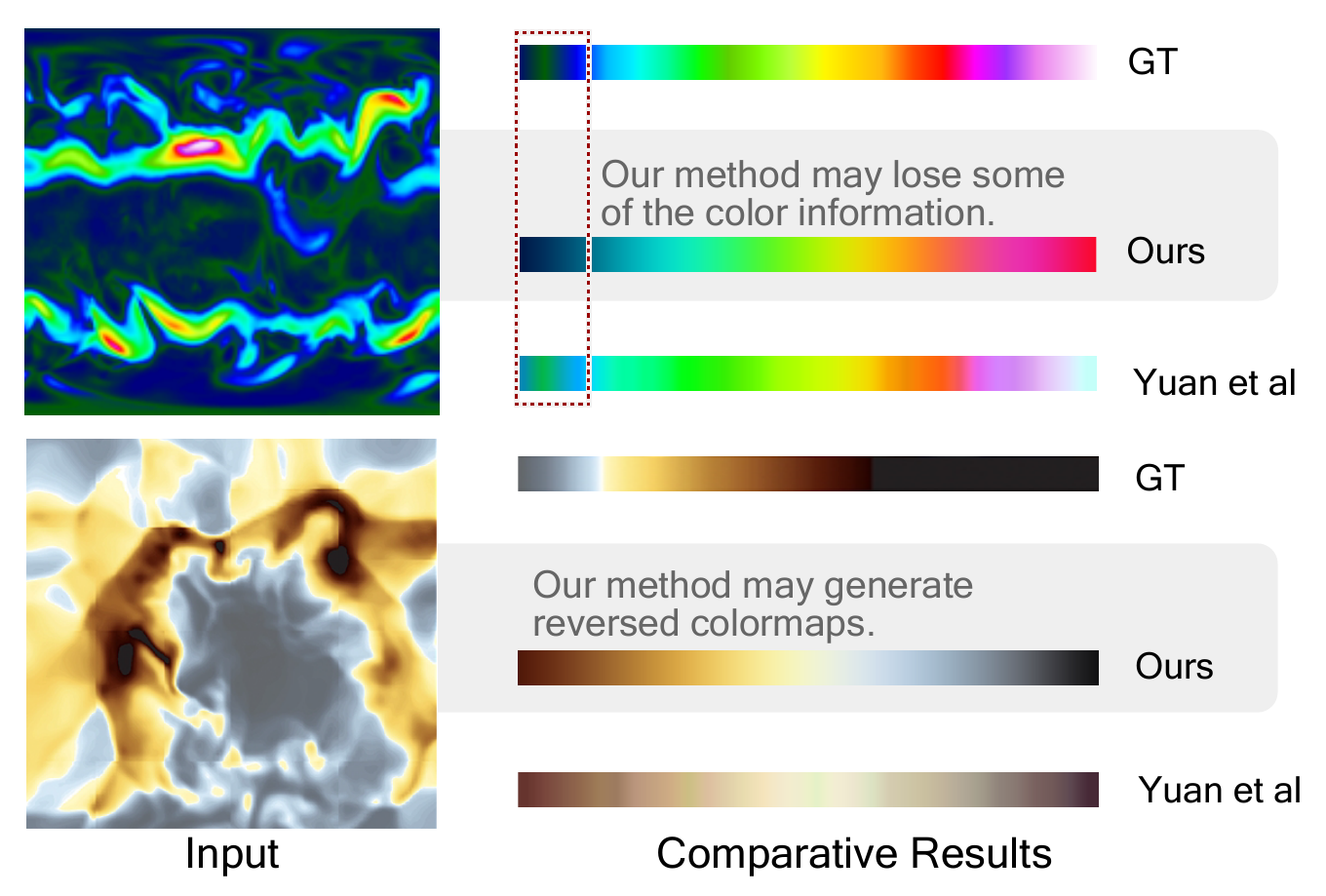}
    \caption{\textbf{Limitations.} Above: for out-of-distribution input visualization, the colormap reconstructed by our method can  at times be poor. Below: the direction of the colormap generated by our method is reversed.}
    \label{fig:failure cases}
\vspace*{-5mm}
\end{figure}

While our approach offers significant potential to unlock hidden information from visualizations without color scales, it exhibits several limitations. 
Our approach is built upon the fundamental assumption of linear mappings between colormaps and data; therefore, its applicability may be limited when dealing with non-linear mappings~\cite{Eisemann11,Thompson13}. 
Our network's performance is currently skewed towards colormaps encountered during training. Less common or user-defined colormaps may lead to less accurate recovery: see~\Cref{fig:failure cases} (above).
Visualizations can be inherently ambiguous. For instance, reversing a colormap and the corresponding data can produce an identical visual representation. 
In such cases, the network may struggle to differentiate between true data variations and colormap ordering, potentially leading to recovered elements that deviate from the intended meaning: see~\Cref{fig:failure cases} (below). 
While the recovered colormap and data are sufficient for course-grained data analysis---such as pattern and distribution exploration---this intrinsic ambiguity makes it challenging to support accurate data analysis. Training and fine-tuning the network can be computationally expensive, requiring significant processing power and resources\delete{. This could limit}\visedit{, which limits the} accessibility for users with limited computational resources. \delete{Currently}\visedit{Additionally}, our network cannot distinguish legend and visualization content, \visedit{which may introduce}\delete{ For a visualization with a legend, both are treated as part of the input, potentially introducing} noise during the colormap \visedit{recovery}\delete{and data prediction} process.

\subsection{Future Work} \visedit{Colormap recovery is a highly challenging problem, and there remains a large body of future work in this direction.} To improve the network's ability to handle non-linear mappings or user-defined colormaps, techniques like data augmentation and domain adaptation can be explored~\cite{li2024comprehensive}. 
Visualizations with inherent ambiguity \visedit{may remain fundamentally unresolved, even when incorporating a large amount of training data, as such data may confuse the network. Addressing these challenges requires incorporating stronger constraints, such as additional perceptual cues~\cite{Bujack18,Schloss19}, uncertainty quantification~\cite{Kamal21,Kong23}, and interactive exploration~\cite{Elmqvist11,Li24} into the learning framework}.
To\delete{ help the network distinguish between true data variations and colormap ordering directions, adding perceptual cues as additional constraints can guide the network toward more robust interpretations.} improve computational efficiency, lightweight network architectures as well as meta-learning techniques, e.g., MAML~\cite{finn2017model}, can be used to reduce the computational burden and inferencing-time optimization.
To improve colormap and data recovery by focusing on relevant visual elements in visualizations with legends, automatic object detection methods~\cite{Ren15} could be introduced to identify and separate legend regions. We also hope to explore\delete{ real-world} applications\delete{ of the network} beyond color design, including integrating the network with scientific visualization tools, data exploration platforms, accessibility solutions for visualizations lacking color legends, and question-answering for visualization interpretation.
 Furthermore, although colormap recovery is beneficial for generating \visedit{accessible} visualizations, it may also pose risks concerning data security, which intrigues us to develop visual encodings for security in the future.

\section{Conclusions} 
We present a novel colormap recovery network that leverages a decoupling-and-reconstruction strategy to accurately predict embedded colormaps and data from visualizations without legends.
Backed by a mathematical analysis and informed by design requirements, the network incorporates a  cubic B-spline colormap representation and four key loss functions for high-fidelity recovery.
Our reconstruction loss function enables self-supervised fine-tuning during inferencing.  
Extensive comparisons and an ablation study demonstrate the superiority of our approach over alternative methods.  
Furthermore, we show the utility of our method with a prototype application with  \emph{visual recoloring} and \emph{data view} functionalities.
These, combined with the core colormap recovery capabilities, unlock the potential of ``in-the-wild'' visualizations---not only facilitating color design, but also potentially improving the accessibility of existing visualizations.



\acknowledgments{
The authors would like to thank Linping Yuan at HKUST and Wei Zeng at HKUST(GZ) for providing valuable discussions and source codes, as well as the anonymous reviewers for their helpful comments. In this study, Qiong Zeng is supported by grants from the National Key R\&D Program of China (No. 2021YFF0704300), the National Natural Science Foundation of China (No. 62372271), and the Taishan Scholars Program (No. tsqn202408291); Manyi Li is supported by the Excellent Young Scientists Fund Program (Overseas) of Shandong Province (No. 2023HWYQ-034); and Yunhai Wang is supported by the National Natural Science Foundation of China (No. 62132017 and No. U2436209), the Natural Science Foundation of Shandong Province (No. ZQ2022JQ32), the Beijing Natural Science Foundation (L247027), and the Fundamental Research Funds for the Central Universities.}

\bibliographystyle{abbrv-doi-hyperref}
\bibliography{coloropt}

\end{document}